\documentclass{article}

\PassOptionsToPackage{numbers,compress}{natbib}
\usepackage[preprint]{neurips_2026}

\usepackage[utf8]{inputenc}
\usepackage[T1]{fontenc}
\usepackage{url}
\usepackage{booktabs}
\usepackage{amsfonts}
\usepackage{amsmath}
\usepackage{amssymb}
\usepackage{nicefrac}
\usepackage{microtype}
\usepackage[table]{xcolor}
\usepackage{graphicx}
\usepackage{wrapfig}
\usepackage{makecell}
\usepackage{tabularx}
\usepackage{array}
\usepackage{multirow}
\usepackage{siunitx}
\usepackage{float}
\usepackage{tikz}
\usepackage[most]{tcolorbox}
\usepackage{pifont}
\usepackage{xspace}
\definecolor{linkred}{RGB}{220,70,140}
\usepackage[colorlinks=true, urlcolor=linkred]{hyperref}
\usepackage{xurl}
\usepackage[nameinlink,noabbrev]{cleveref}
\usepackage{caption}
\definecolor{paperink}{HTML}{2D3748}
\definecolor{backred}{RGB}{255,190,190}
\definecolor{backblue}{RGB}{210,230,250}
\definecolor{myred}{HTML}{EA4335}
\definecolor{mygreen}{HTML}{34A853}
\definecolor{lightblue}{rgb}{0.529, 0.808, 0.922}
\definecolor{MI}{RGB}{231,236,222}
\definecolor{MA}{RGB}{234,240,182}
\definecolor{MR}{RGB}{225,234,207}
\definecolor{TR}{RGB}{213,227,186}
\definecolor{LR}{RGB}{254,208,215}
\definecolor{BR}{RGB}{242,189,193}
\definecolor{HR}{RGB}{254,202,204}
\definecolor{PR}{RGB}{248,191,179}
\definecolor{GR}{RGB}{187,222,247}
\definecolor{AR}{RGB}{179,211,248}
\definecolor{CR}{RGB}{253,212,191}
\definecolor{RR}{RGB}{254,205,161}

\newcommand{\dataset}{\datasettitle\xspace}
\newcommand{\datasettitle}{\textit{MathGen}}
\newcommand{\best}[1]{\cellcolor{backred}\textbf{#1}}
\newcommand{\high}[1]{\cellcolor{backblue}#1}

\newcommand{\greenyes}{\textcolor{green}{\ding{51}}}
\newcommand{\redno}{\textcolor{red}{\ding{55}}}

\newcolumntype{Y}{>{\centering\arraybackslash}X}
\newcolumntype{C}[1]{>{\centering\arraybackslash}m{#1}}
\newcolumntype{L}[1]{>{\raggedright\arraybackslash}m{#1}}

\newtcolorbox[auto counter]{myfinding}[1][]{
  enhanced,
  breakable,
  colframe=lightblue,
  colback=lightblue!30!white,
  sharp corners,
  boxsep=0pt,
  left=5pt,
  right=5pt,
  top=6pt,
  bottom=6pt,
  boxrule=0pt,
  leftrule=4pt,
  before skip=10pt,
  after skip=0pt,
  before upper={\textbf{Finding \thetcbcounter: } },
  #1
}

\title{
\begin{minipage}{.08\textwidth}
\centering
\vspace{-4pt}
\includegraphics[width=\linewidth]{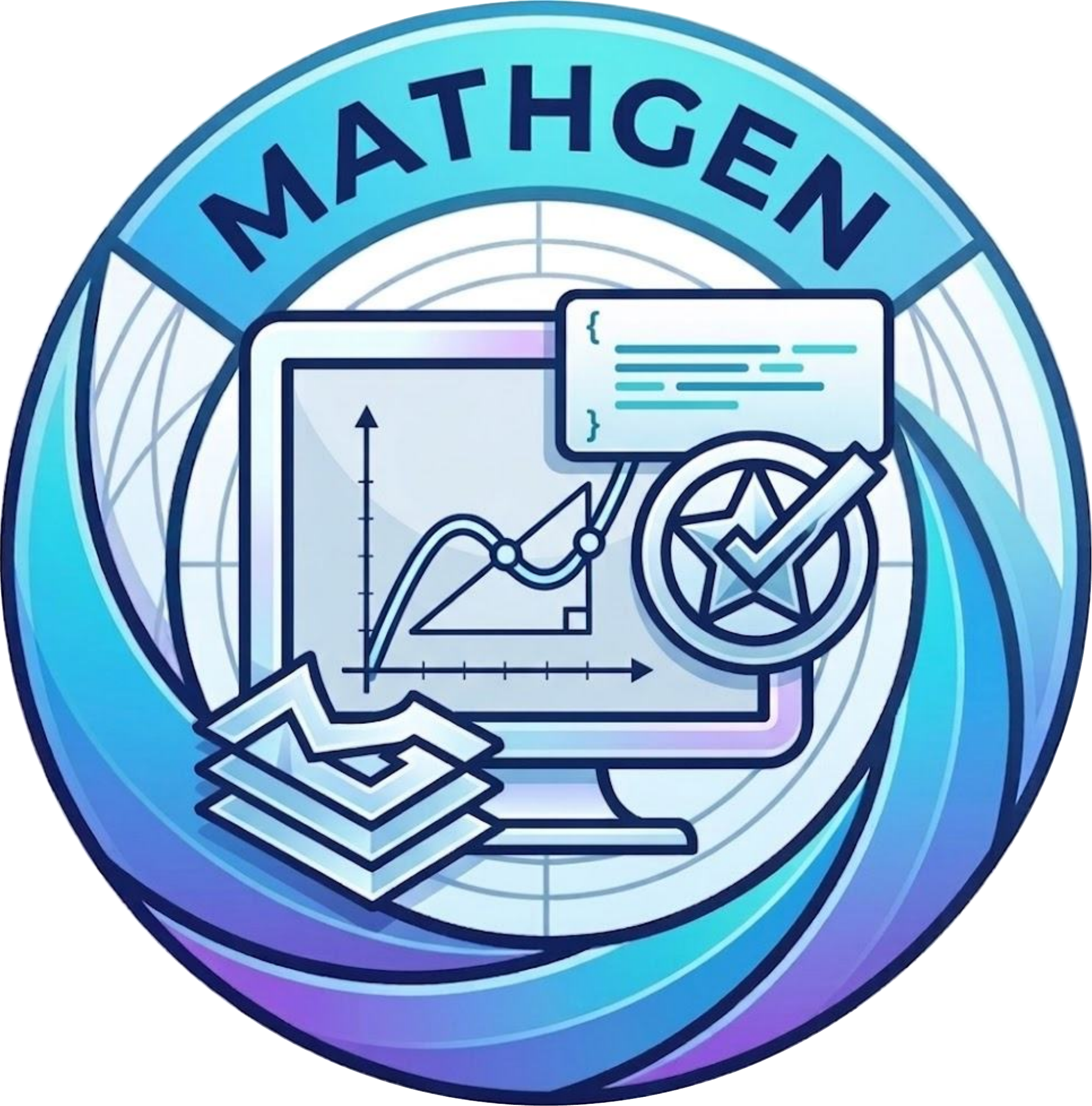} 
\end{minipage}
\datasettitle: Revealing the Illusion of Mathematical Competence through Text-to-Image Generation}
\author{Ruiyao Liu$^{1,*}$, Hui Shen$^{2,}$\thanks{denotes co-first authors. $\dagger$ denotes corresponding authors} , Ping Zhang$^{3,*}$, Yunta Hsieh$^{2}$,\vspace{0.1cm}\\ \textbf{Yifan Zhang$^2$, Jing Xu$^4$, Qi Han$^7$, Junchen Li$^{5}$, Jiawei Lu$^{6}$,}\vspace{0.1cm}\\ \textbf{Jianing Ma$^7$, Jiaqi Mo$^6$, Sicheng Chen$^3$, Zhen Zhang$^{8}$, Zhongwei Wan$^{3}$,}\vspace{0.1cm}\\ \textbf{Jing Xiong$^{9}$, Xin Wang$^{3}$, Ziyuan Liu$^{10}$, Hangrui Cao$^{11}$, Ngai Wong$^{9,\dagger}$}\vspace{0.3cm}\\ 
  $^1$University of Pennsylvania, 
  $^2$University of Michigan, 
  $^3$The Ohio State University,
  $^4$USTC,\\
  $^5$City University of Hong Kong,
  $^6$University of Wisconsin,
  $^7$Independent,\\
  $^8$UCSB,
  $^9$University of Hong Kong,
  $^{10}$Peking University,
  $^{11}$Carnegie Mellon University
}
\begin{document}

\maketitle

\begin{abstract}
Modern generative models have demonstrated the ability to solve challenging mathematical problems. In many real-world settings, however, mathematical solutions must be expressed visually through diagrams, plots, geometric constructions, and structured symbolic layouts, where correctness depends on precise visual composition. This naturally raises the question of~\emph{whether generative models can still do so when the answer must be rendered visually rather than written in text?} To study this problem, we introduce \textbf{\dataset}, a rigorous benchmark of $420$ problems spanning seven core domains. Each problem is evaluated under a \emph{Script-as-a-Judge} protocol with problem-specific verification criteria implemented through reusable executable scripts for deterministic and reproducible evaluation. Experiments on representative open-source and proprietary text-to-image models show that mathematical fidelity remains a major bottleneck: even the best closed-source model reaches only $53.7\%$ overall accuracy, while open-source models achieve just $\sim 1$--$11\%$, often near $0\%$ on structured tasks, particularly those requiring precise geometric and functional rendering. Overall, current T2I models remain far from reliable at even elementary mathematical visual generation. More details are available on our project page: \href{https://mathgen-t2i.github.io/}{mathgen-t2i.github.io}.
\end{abstract}
\section{Introduction}
Recent progress in generative models has substantially improved mathematical reasoning in text. Large language models can now solve a wide range of challenging problems, including competition-level questions, standardized exam problems, and structured symbolic reasoning tasks~\citep{lu2023mathvista, wang2025mv, wang2024measuring, feng2025mathreal}. In parallel, visual generation has advanced rapidly through diffusion models~\citep{saharia2022photorealistic, MaGABVX24, EsserKBEMSLLSBP24} and autoregressive paradigms~\citep{yu2022scaling}, enabling state-of-the-art text-to-image (T2I) systems to produce high-fidelity images from natural language descriptions. This raises a natural question: does mathematical competence persist when solutions must be rendered visually rather than textually?

\begin{figure}[h]
    \centering
    \includegraphics[width=1\linewidth]{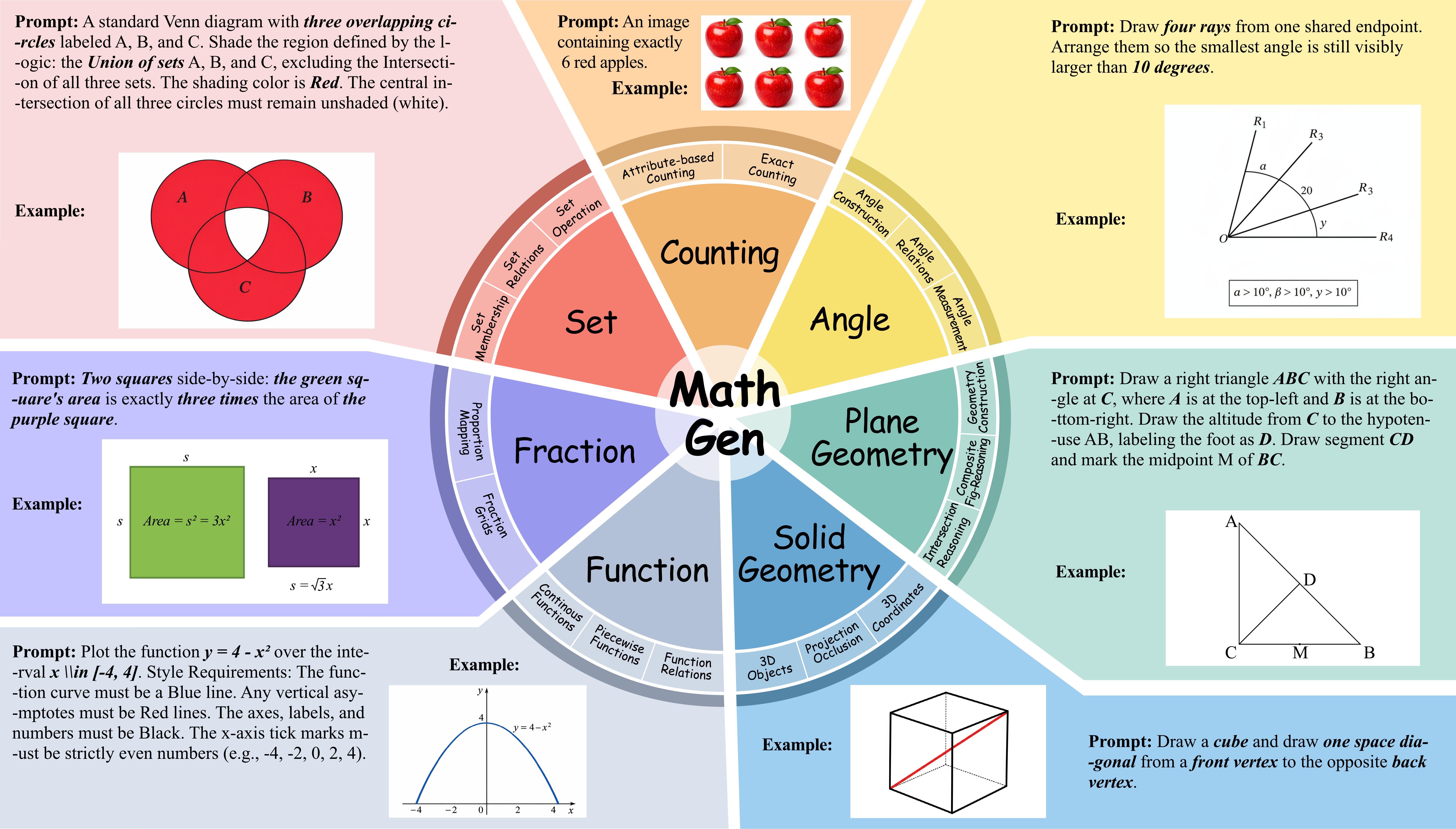}  
    \caption{\small Task taxonomy of \dataset.
    \dataset covers seven fundamental mathematical domains. Example prompts and reference illustrations are shown to provide an overview of the mathematical concepts evaluated in each domain. They highlight the diverse forms of numerical, geometric, and structural constraints that generative models are required to express, and illustrate the types of visual outcomes expected under correct mathematical interpretation.
    }
    \label{fig:overview}
    \vspace{-2mm}
\end{figure}

Answering this question requires evaluating whether generative models can reliably preserve numerical, geometric, and relational constraints in visual form. As illustrated in Fig.~\ref{fig:overview}, mathematical image generation spans multiple domains with distinct structural requirements, making correctness difficult to assess using standard perceptual criteria alone. Even small deviations from these constraints can invalidate the intended interpretation of a generated figure.

\begin{wrapfigure}{r}{0.5\textwidth}
    \centering
    \includegraphics[width=0.5\textwidth]{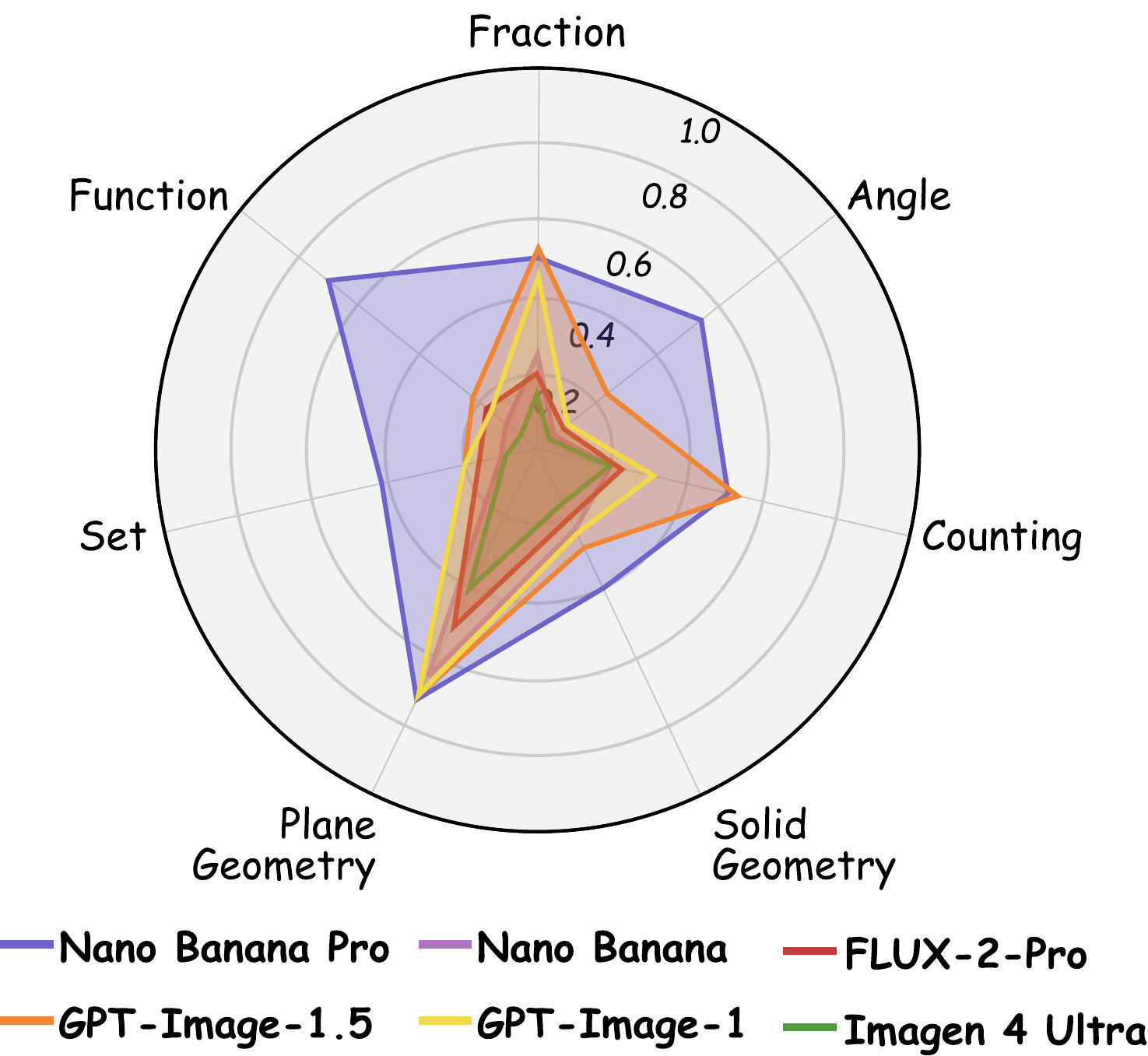}
    \caption{\small Performance comparison. The chart shows the accuracy of representative open-source and closed-source text-to-image models on each \dataset domain.
    }
    \label{fig:radar}
    \vspace{-4mm}
\end{wrapfigure}
Despite its importance, mathematical correctness remains poorly measured in existing evaluation pipelines. Most current T2I benchmarks emphasize semantic alignment, compositionality, or visual realism~\citep{huang2023t2i, li2025easier, wang2025genexam}, while only a small number of recent efforts consider mathematics-related visual generation settings~\citep{DBLP:conf/acl/WangRWS25}. In practice, evaluation is often delegated to vision-language models (VLMs), whose judgments can be unreliable for fine-grained structural verification~\citep{Tong0Z0LX24}. As a result, current benchmarks provide limited insight into whether generated images satisfy the underlying mathematical constraints.

Moreover, widely used automatic metrics such as CLIP-score~\citep{hessel2021clipscore} and FID~\citep{heusel2017gans} measure semantic similarity or distributional realism rather than exact structural validity. These metrics are therefore insufficient for assessing mathematical generation tasks that require precise verification of mathematical constraints.

To address this gap, we propose a tool-based evaluation protocol. For each problem, we define problem-specific verification criteria and evaluate them using reusable executable scripts and computer vision modules that extract the relevant structure from a generated image and deterministically check whether the mathematical requirements are satisfied. This \emph{Script-as-a-Judge} framework enables deterministic, reproducible, and fine-grained evaluation of mathematical correctness in text-to-image generation.

Building on this protocol, we introduce \textbf{\dataset}, a benchmark for assessing mathematical generation in T2I models. \dataset contains $420$ problems spanning seven core mathematical dimensions: \textit{Counting}, \textit{Fractions}, \textit{Angles}, \textit{Functions}, \textit{Plane Geometry}, \textit{Solid Geometry}, and \textit{Sets}. The Clean-Scene set contains $350$ problems, with $50$ problems per domain. We additionally construct $70$ paired Open-Scene problems, $10$ per domain, from selected Clean-Scene problems. Within each pair, the numerical values, mathematical relations, required answer, and verification criteria are fixed; only the visual realization is grounded in a richer real-world context. This paired design separates failures in executing the mathematical constraint from additional interference introduced by realistic scene composition.

On \dataset, we find that mathematical fidelity remains a major weakness of current T2I models: most evaluated open-source models remain below $10\%$ overall accuracy, with FLUX-2 reaching the highest overall accuracy of $11.1\%$, and several structured domains (e.g., function/angle/set) often near $0\%$. Closed-source models perform substantially better, with Nano Banana Pro reaching $53.7\%$ overall accuracy and GPT-Image-1.5 reaching $39.1\%$. Overall, these results highlight a clear gap between visually plausible generation and constraint-faithful mathematical rendering, motivating systematic and reliable evaluation.

Our contributions can be summarized as follows:
\begin{enumerate}
    \item We present \textbf{\dataset}, the first comprehensive benchmark dedicated to testing mathematical correctness for text-to-image models across Clean-Scene and Open-Scene settings.
    \item We propose a scripted, tool-based evaluation framework that provides deterministic, fine-grained correctness checks for mathematical image generation.
    \item We evaluate representative open-source and closed-source T2I models on \dataset, revealing key failure modes and the limitations of current evaluation practices.
\end{enumerate}

\section{Related Work}
\begin{table}[t]
\centering
\setlength{\tabcolsep}{3pt}
\renewcommand{\arraystretch}{1.2}

\caption{\small\textbf{Comparison with existing benchmarks.}
Prior datasets mainly evaluate visual mathematical understanding,
while \dataset focuses on mathematical image generation with
script-based verification.}
\vspace{2mm}
\scriptsize
\begin{tabular}{p{4cm}p{5.5cm}>{\centering\arraybackslash}m{2.5cm}}
\toprule
\textbf{Benchmark} & \textbf{Theme} & \textbf{Deterministic Eval} \\
\midrule
Geometry3K~\citep{lu2021inter} & Geometry question answering & \redno \\
ChartQA~\citep{masry2022chartqa} & Chart reasoning from visual plots & \redno \\
MathVista~\citep{lu2023mathvista} & Visual mathematical reasoning tasks & \redno \\
T2I-CompBench~\citep{huang2023t2i} & Compositional text-to-image generation & \redno \\
GenExam~\citep{wang2025genexam} & Multidisciplinary generation tasks & \redno \\
Math2Visual~\citep{DBLP:conf/acl/WangRWS25} & Puzzle-solving & \redno \\
\midrule
\textbf{\dataset} & \textbf{Mathematical text-to-image generation} & \greenyes \\
\bottomrule
\end{tabular}

\end{table}

\noindent
\textbf{Mathematical Reasoning in Vision.}
While visual mathematical reasoning has recently garnered significant attention, the community's focus remains overwhelmingly anchored in the \textit{understanding} capabilities of Vision-Language Models, spanning geometry parsing (e.g., Geometry3K~\cite{lu2021inter}), chart reasoning (e.g., ChartQA~\cite{masry2022chartqa}), and mathematical evaluations (e.g., MathVista~\cite{lu2023mathvista}, MATH-Vision~\cite{wang2024measuring}). Unlike VLMs that extract logic from pixels, T2I models face the distinct and complementary challenge of translating abstract mathematical constraints, such as exact ratios, precise intersections, and topological correctness, directly into pixel space. This generative setting remains a critical blind spot in the current landscape.

\noindent
\textbf{Text-to-Image Evaluation.}
Early evaluation of text-to-image (T2I) models mainly relied on perceptual metrics such as Fr\'echet Inception Distance (FID)~\cite{heusel2017gans} and CLIP-Score~\cite{hessel2021clipscore} to assess fidelity and text--image alignment. Recent benchmarks move toward more targeted diagnostics. T2I-CompBench~\cite{huang2023t2i} evaluates compositional generation across attributes, relations, and counting, while T2I-COREBENCH~\cite{li2025easier} further unifies composition and reasoning, revealing reasoning as a key bottleneck. In parallel, GenExam~\cite{wang2025genexam} formulates image generation as multidisciplinary exams with strict scoring, emphasizing semantic correctness over visual plausibility. To enable scalable evaluation, many works adopt VLMs as automatic judges. However, VLM-based judges are themselves prone to hallucination and overconfidence. Despite this progress, existing benchmarks still prioritize perceptual realism and semantic consistency, rarely verifying mathematical correctness. \dataset\ addresses this gap by explicitly requiring precise adherence to numerical and geometric truths in generated images. 
\section{The \dataset Benchmark}

\subsection{Overview of \dataset}

We introduce \dataset, a benchmark designed to systematically evaluate the mathematical capabilities of text-to-image models. 
\dataset comprises $420$ curated problems spanning seven core mathematical domains: \textit{Counting}, \textit{Angles}, \textit{Fractions and Ratios}, \textit{Plane Geometry}, \textit{Functions}, \textit{Sets}, and \textit{Solid Geometry}. The Clean-Scene set contains $350$ problems, with $50$ problems per domain, and the Open-Scene set contains $70$ paired problems, with $10$ per domain, constructed from selected Clean-Scene problems.
Each problem describes a mathematical concept that must be translated into a visually valid representation. Representative question examples from each domain are illustrated in Figure~\ref{fig:overview}. All questions in our benchmark were manually collected by graduate students in STEM fields. Through its carefully curated structure and broad coverage of mathematical domains, \dataset provides a systematic resource for benchmarking and advancing the mathematical generation capabilities of foundation T2I models.

\begin{figure}
    \centering
    \includegraphics[width=1\linewidth]{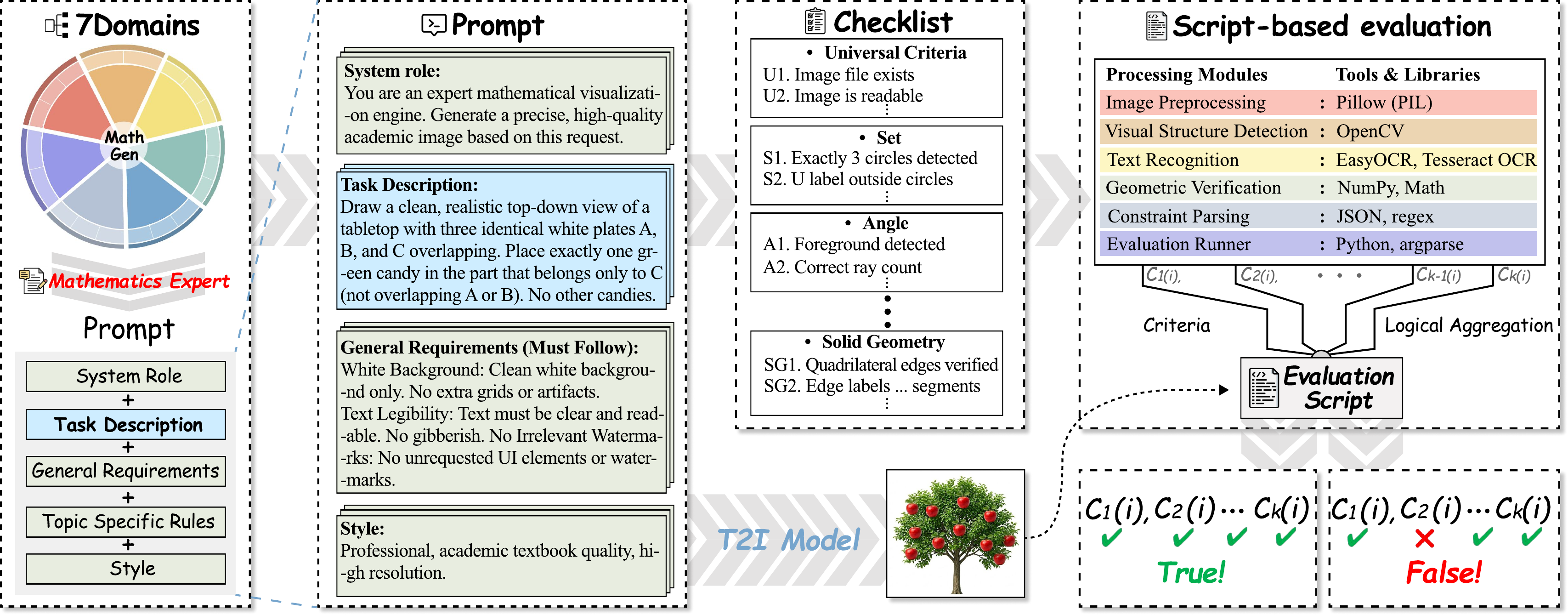}
    \caption{\small Overview of the \dataset benchmark and evaluation pipeline.
    \dataset evaluates text-to-image models on seven mathematical domains using structured prompts and automatic verification.
    Generated images are validated against domain-specific structural, geometric, and logical constraints.
    Each criterion $C_i$ is checked independently, and the final correctness is determined through logical aggregation.
    }
    \label{fig:pipeline}
\end{figure}

\subsection{Deterministic Evaluation}
\label{sec:deterministic_eval}

As shown in Figure~\ref{fig:pipeline}, to reduce the subjectivity and variability associated with VLM-based evaluation, \dataset employs a deterministic evaluation protocol. We develop a collection of rule-based verification scripts and reusable processing modules that evaluate problem-specific mathematical constraints, primarily using classical computer vision tools.

Expert annotators translate abstract mathematical constraints into programmable, pixel-level verifications, such as contour analysis for shape extraction, Hough transforms for line intersection detection, and region-based topological checks for set reasoning.

For each problem $p$, we define a set of deterministic constraints $\mathcal{C}(p)$ describing the mathematical conditions that a generated image $I$ must satisfy. These constraints are evaluated using executable Python scripts and reusable verification modules based on classical computer vision techniques, including contour detection, OCR, and pixel-level color analysis. A sample is considered correct only when all constraints are simultaneously satisfied. 

This evaluation design offers three key advantages: 
(1) \textbf{determinism}, as the evaluation logic is explicitly defined and reproducible; 
(2) \textbf{transparency}, since each decision can be traced to explicit and measurable visual criteria; and 
(3) \textbf{reproducibility}, enabling the entire evaluation pipeline to be executed locally without reliance on a VLM as judge.
Moreover, encoding the evaluation criteria as explicit executable scripts makes the verifiers easy to inspect, refine, and continuously improve through community feedback.

To further assess the reliability of our script-based evaluation, five graduate student volunteers with STEM backgrounds participated in the human evaluation, with the script-based judgments achieving an overall agreement of $95.4\%$ with human annotations. We additionally manually audit cases with intentionally violated constraints to verify that incorrect generations are consistently detected.

For the counting domain, which requires instance-level localization before numerical verification, we employ a lightweight RT-DETR detector (R50-VD backbone), and a sample is considered correct only when the detected number of target objects exactly matches the required quantity.

\subsection{Task Taxonomy and Problem Design}

To comprehensively evaluate mathematical generation capability, we design tasks from two complementary perspectives: clean-scene math and open-scene math. The former focuses on controlled diagrammatic reasoning under explicit mathematical constraints, while the latter examines whether models can preserve mathematical correctness in visually complex real-world scenes. 

\subsubsection{Clean-scene Math Design}

We organize clean-scene math into seven domains that capture fundamental forms of mathematical reasoning.

\paragraph{Counting.}
Counting evaluates whether models can generate scenes satisfying strict numerical constraints. This domain includes \textit{Exact Counting}, which requires generating an exact number of target objects, and \textit{Attribute-based Counting}, in which only objects satisfying specified attribute conditions (e.g., color or shape) should be counted while preserving the required quantity.

\paragraph{Angles.}
Angle tasks assess the ability to construct geometrically valid angular structures. These tasks include \textit{Angle Measurement}, which requires depicting angles with the correct magnitude or type; \textit{Angle Relations}, which represents relationships among multiple angles, such as equality, complementarity, or ordering; and \textit{Angle Construction}, which requires generating an angle that satisfies a given geometric condition.

\paragraph{Fractions and Ratios.}
This domain evaluates proportional reasoning and fractional representation. Tasks include \textit{Proportion Mapping}, which visually represents continuous proportional relationships, and \textit{Fraction Grids}, which divide objects or regions into discrete partitions corresponding to fractional values.

\paragraph{Plane Geometry.}
Plane geometry tasks examine spatial reasoning over two-dimensional geometric constructions. These include \textit{Intersection Reasoning}, which requires correctly constructing intersections among lines, curves, or shapes; \textit{Composite Figure Reasoning}, which requires generating structures composed of multiple geometric primitives; and \textit{Geometric Construction}, which requires generating or completing planar figures under explicit geometric constraints.

\paragraph{Functions.}
Function tasks evaluate whether models can accurately visualize mathematical functions. This domain includes \textit{Continuous Functions}, which require plotting continuous curves on coordinate axes; \textit{Piecewise Functions}, which represent functions defined by multiple segments or rules; and \textit{Function Relations}, which capture relations among multiple functions, such as intersection, ordering, or alignment.

\paragraph{Sets.}
Set tasks measure logical reasoning through spatial and symbolic set representations. These tasks include \textit{Set Operations}, which depict operations such as union, intersection, difference, or complement; \textit{Set Relations}, which represent structural relations such as subset, overlap, equality, or disjointness; and \textit{Set Membership}, which places elements in the correct set regions according to membership constraints.

\paragraph{Solid Geometry.}
Solid geometry tasks assess spatial reasoning in three-dimensional settings. These include \textit{3D Shape Recognition}, which requires generating geometrically correct three-dimensional objects; \textit{Spatial Position \& Coordinates}, which evaluates whether objects or points are placed consistently according to spatial or coordinate constraints; and \textit{Projection, Visibility \& Occlusion}, which evaluates projection structure together with visible and hidden-part relations.

\subsubsection{Open-scene Math Design}
\begin{table*}[t]
\centering
\caption{\textbf{Paired Clean-Scene/Open-Scene comparison} on $70$ matched mathematical instances ($10$ per domain). Each pair preserves the same numerical values, mathematical relations, required answer, and evaluation criteria; only the scene realization and necessary lexical grounding differ. Both rows report overall accuracy (\%), computed as the number of correct generations divided by the corresponding $70$ prompts.}
\vspace{-2mm}

\fontsize{6pt}{7pt}\selectfont
\setlength{\tabcolsep}{3.5pt}
\setlength{\aboverulesep}{0.5ex}
\setlength{\belowrulesep}{0.5ex}
\renewcommand{\arraystretch}{1.35}
\setlength{\extrarowheight}{0.4pt}

\begin{tabularx}{\linewidth}{
>{\centering\arraybackslash}m{0.13\linewidth}|
*{4}{>{\centering\arraybackslash}X|}%
>{\centering\arraybackslash}X
}
\toprule

\textbf{Set} &
\makecell{\textbf{SD-3}\\\textbf{Medium}} &
\makecell{\textbf{SD-3.5}\\\textbf{Medium}} &
\makecell{\textbf{SD-3.5}\\\textbf{Large}} &
\makecell{\textbf{FLUX-2}} &
\makecell{\textbf{FLUX-2}\\\textbf{Pro}} \\
\midrule

\textbf{Clean-Scene} & 2.9 & 2.9 & 4.3 & 12.9 & 30.0 \\
\rowcolor{black!6}
\textbf{Open-Scene}  & 10.0 & 8.6 & 5.7 & 7.1 & 12.9 \\

\specialrule{0.9pt}{0.4ex}{0.4ex}

\multicolumn{1}{c|}{} &
\makecell{\textbf{FLUX}\\\textbf{Kontext-Pro}} &
\makecell{\textbf{PixArt}\\\textbf{$\Sigma$}} &
\makecell{\textbf{PixArt}\\\textbf{XL-2}} &
\makecell{\textbf{HiDream}\\\textbf{I1}} &
\makecell{\textbf{Qwen}\\\textbf{Image}} \\
\midrule

\textbf{Clean-Scene} & 10.0 & 4.3 & 4.3 & 2.9 & 12.9 \\
\rowcolor{black!6}
\textbf{Open-Scene}  & 10.0 & 2.9 & 4.3 & 1.4 & 10.0 \\

\specialrule{0.9pt}{0pt}{0pt}

\multicolumn{1}{c|}{} &
\makecell{\textbf{Infinity}\\\textbf{8B}} &
\makecell{\textbf{GoT-R1}\\\textbf{7B}} &
\makecell{\textbf{BAGEL}} &
\makecell{\textbf{show-o2}\\\textbf{1.5B}} &
\makecell{\textbf{show-o2}\\\textbf{7B}} \\
\midrule

\textbf{Clean-Scene} & 5.7 & 4.3 & 4.3 & 4.3 & 0.0 \\
\rowcolor{black!6}
\textbf{Open-Scene}  & 4.3 & 4.3 & 10.0 & 4.3 & 2.9 \\

\specialrule{0.9pt}{0pt}{0pt}

\multicolumn{1}{c|}{} &
\makecell{\textbf{Janus}\\\textbf{Pro-1B}} &
\makecell{\textbf{Janus}\\\textbf{Pro-7B}} &
\makecell{\textbf{BLIP3o}\\\textbf{4B}} &
\makecell{\textbf{BLIP3o}\\\textbf{8B}} &
\makecell{\textbf{OmniGen2}\\\textbf{7B}} \\
\midrule

\textbf{Clean-Scene} & 2.9 & 4.3 & 8.6 & 5.7 & 1.4 \\
\rowcolor{black!6}
\textbf{Open-Scene}  & 1.4 & 1.4 & 1.4 & 4.3 & 10.0 \\

\specialrule{0.9pt}{0pt}{0pt}

\multicolumn{1}{c|}{} &
\makecell{\textbf{Seedream}\\\textbf{3.0}} &
\makecell{\textbf{Seedream}\\\textbf{4.0}} &
\makecell{\textbf{Ideogram}\\\textbf{v3 Turbo}} &
\makecell{\textbf{Nano}\\\textbf{Banana}} &
\makecell{\textbf{Nano}\\\textbf{Banana Pro}} \\
\midrule

\textbf{Clean-Scene} & 11.4 & 17.1 & 7.1 & 25.7 & 42.9 \\
\rowcolor{black!6}
\textbf{Open-Scene}  & 10.0 & 2.9 & 5.7 & 21.4 & 32.9 \\

\specialrule{0.9pt}{0pt}{0pt}

\multicolumn{1}{c|}{} &
\makecell{\textbf{Imagen}\\\textbf{4}} &
\makecell{\textbf{Imagen}\\\textbf{4 Ultra}} &
\makecell{\textbf{GPT-Image}\\\textbf{1}} &
\makecell{\textbf{GPT-Image}\\\textbf{1.5}} &
\makecell{\textbf{Z-Image}\\\textbf{Turbo}} \\
\midrule

\textbf{Clean-Scene} & 2.9 & 25.7 & 32.9 & 38.6 & 5.7 \\
\rowcolor{black!6}
\textbf{Open-Scene}  & 7.1 & 20.0 & 25.7 & 18.6 & 0.0 \\

\bottomrule
\end{tabularx}

\vspace{-2mm}
\label{table:clean_open}
\end{table*}

To evaluate mathematical reasoning under different levels of visual complexity, we introduce the open-scene math set.
This set contains realistic scenes with richer backgrounds and more complex compositions. Each Open-Scene problem is paired with a corresponding Clean-Scene problem while preserving the same underlying numerical values, mathematical relations, required answer, and verification criteria. For reliable script-based evaluation, the Open-Scene prompts are lightly refined to encourage visually interpretable layouts, such as clear viewpoints and visible key structures, without changing the underlying mathematical requirements. This design allows us to measure the additional difficulty introduced by realistic scene composition while preserving the same mathematical constraints.

\begin{table*}
\centering
\caption{Main results on the $350$-problem Clean-Scene set of \dataset. We report accuracy across seven mathematical domains, with $50$ problems per domain.
Closed-source models generally achieve the strongest overall performance, although performance remains far from perfect in several domains, particularly set and solid-geometry tasks.
Best results in each column are highlighted in \best{red} and the second-best results in \high{blue}.
}
\setlength{\tabcolsep}{4.2pt}
\renewcommand{\arraystretch}{1.0}
\small
\resizebox{\linewidth}{!}{
\begin{tabular}{lcccccccc}
\toprule
\textbf{Model}
& \textbf{counting}
& \textbf{angle}
& \textbf{fraction}
& \textbf{function}
& \textbf{plane}
& \textbf{set}
& \textbf{solid}
& \cellcolor[HTML]{E2E2E2}\textbf{Overall} \\
\midrule

\rowcolor[HTML]{F3F5F7}
\multicolumn{9}{c}{\textit{\textbf{Diffusion Models}}} \\
\midrule

SD-3-Medium     & 0.0 & 0.0 & 0.0 & 0.0 & 6.0 & 0.0 & 6.0 & \cellcolor[HTML]{E2E2E2}1.7 \\
SD-3.5-Medium   & 4.0 & 0.0 & 0.0 & 0.0 & 12.0 & 2.0 & 6.0 & \cellcolor[HTML]{E2E2E2}3.4 \\
SD-3.5-Large    & 12.0 & 0.0 & 2.0 & 0.0 & 12.0 & 2.0 & 4.0 & \cellcolor[HTML]{E2E2E2}4.6 \\
FLUX-2          & 8.0 & 2.0 & 8.0 & 2.0 & 42.0 & 8.0 & 8.0 & \cellcolor[HTML]{E2E2E2}11.1 \\
PixArt-$\Sigma$ & 10.0 & 0.0 & 2.0 & 0.0 & 12.0 & 2.0 & 8.0 & \cellcolor[HTML]{E2E2E2}4.9 \\
PixArt-XL-2     & 0.0 & 0.0 & 0.0 & 0.0 & 14.0 & 0.0 & 4.0 & \cellcolor[HTML]{E2E2E2}2.6 \\
HiDream-I1      & 6.0 & 0.0 & 0.0 & 2.0 & 4.0 & 2.0 & 6.0 & \cellcolor[HTML]{E2E2E2}2.9 \\
Qwen-Image      & 22.0 & 0.0 & 8.0 & 2.0 & 24.0 & 4.0 & 6.0 & \cellcolor[HTML]{E2E2E2}9.4 \\
Z-Image-Turbo   & 8.0 & 0.0 & 8.0 & 0.0 & 16.0 & 2.0 & 14.0 & \cellcolor[HTML]{E2E2E2}6.9 \\

\addlinespace[1mm]
\rowcolor[HTML]{F3F5F7}
\multicolumn{9}{c}{\textit{\textbf{Autoregressive Models}}} \\
\midrule

Infinity-8B     & 6.0 & 0.0 & 4.0 & 0.0 & 18.0 & 2.0 & 8.0 & \cellcolor[HTML]{E2E2E2}5.4 \\
GoT-R1-7B       & 8.0 & 0.0 & 0.0 & 0.0 & 16.0 & 2.0 & 2.0 & \cellcolor[HTML]{E2E2E2}4.0 \\

\addlinespace[1mm]
\rowcolor[HTML]{F3F5F7}
\multicolumn{9}{c}{\textit{\textbf{Unified Models}}} \\
\midrule

BAGEL           & 4.0 & 0.0 & 0.0 & 0.0 & 14.0 & 0.0 & 2.0 & \cellcolor[HTML]{E2E2E2}2.9 \\
show-o2-1.5B    & 0.0 & 0.0 & 0.0 & 4.0 & 18.0 & 0.0 & 2.0 & \cellcolor[HTML]{E2E2E2}3.4 \\
show-o2-7B      & 0.0 & 0.0 & 0.0 & 4.0 & 10.0 & 0.0 & 8.0 & \cellcolor[HTML]{E2E2E2}3.1 \\
Janus-Pro-1B    & 0.0 & 0.0 & 0.0 & 0.0 & 14.0 & 0.0 & 2.0 & \cellcolor[HTML]{E2E2E2}2.3 \\
Janus-Pro-7B    & 0.0 & 0.0 & 0.0 & 0.0 & 12.0 & 2.0 & 6.0 & \cellcolor[HTML]{E2E2E2}2.9 \\
BLIP3o-4B       & 4.0 & 0.0 & 2.0 & 0.0 & 14.0 & 2.0 & 8.0 & \cellcolor[HTML]{E2E2E2}4.3 \\
BLIP3o-8B       & 6.0 & 0.0 & 2.0 & 0.0 & 14.0 & 2.0 & 4.0 & \cellcolor[HTML]{E2E2E2}4.0 \\
OmniGen2-7B     & 4.0 & 0.0 & 2.0 & 0.0 & 12.0 & 2.0 & 8.0 & \cellcolor[HTML]{E2E2E2}4.0 \\

\addlinespace[1mm]
\rowcolor[HTML]{F3F5F7}
\multicolumn{9}{c}{\textit{\textbf{Closed-Source Models}}} \\
\midrule

FLUX-2-Pro       & 22.0 & 10.0 & 20.0 & 18.0 & 54.0 & 16.0 & 20.0 & \cellcolor[HTML]{E2E2E2}22.9 \\
FLUX-Kontext-Pro & 10.0 & 0.0 & 10.0 & 4.0 & 18.0 & 6.0 & 4.0 & \cellcolor[HTML]{E2E2E2}7.4 \\
Seedream 3.0     & 14.0 & 0.0 & 2.0 & 0.0 & 24.0 & 6.0 & 8.0 & \cellcolor[HTML]{E2E2E2}7.7 \\
Seedream 4.0     & 20.0 & 0.0 & 6.0 & 10.0 & 36.0 & 6.0 & 14.0 & \cellcolor[HTML]{E2E2E2}13.1 \\
Ideogram v3 Turbo& 10.0 & 2.0 & 0.0 & 0.0 & 20.0 & 2.0 & 8.0 & \cellcolor[HTML]{E2E2E2}6.0 \\
Nano Banana      & 20.0 & 8.0 & 24.0 & 10.0 & 64.0 & 10.0 & 24.0 & \cellcolor[HTML]{E2E2E2}22.9 \\
Nano Banana Pro  & \high{48.0} & \best{54.0} & \high{50.0} & \best{70.0} & \best{72.0} & \best{42.0} & \best{40.0} & \cellcolor[HTML]{E2E2E2}\best{53.7} \\
Imagen 4         & 12.0 & 2.0 & 2.0 & 2.0 & 12.0 & 0.0 & 12.0 & \cellcolor[HTML]{E2E2E2}6.0 \\
Imagen 4 Ultra   & 20.0 & 6.0 & 16.0 & 4.0 & 42.0 & 8.0 & 16.0 & \cellcolor[HTML]{E2E2E2}16.0 \\
GPT-Image-1      & 32.0 & 12.0 & 44.0 & 16.0 & 68.0 & \high{20.0} & 24.0 & \cellcolor[HTML]{E2E2E2}30.9 \\
GPT-Image-1.5    & \best{56.0} & \high{24.0} & \best{54.0} & \high{22.0} & \high{70.0} & \high{20.0} & \high{28.0} & \cellcolor[HTML]{E2E2E2}\high{39.1} \\
\bottomrule
\end{tabular}
}
\label{table:main_results}
\end{table*}
\section{Experiments}

\subsection{Experimental Setup}

\textbf{Evaluated Models.}
\vspace{2mm}
We evaluate a diverse set of recent text-to-image models spanning three major architectural paradigms: diffusion models, autoregressive models, and unified multimodal generation models.
The evaluated diffusion models include Stable Diffusion variants (SD-3-Medium, SD-3.5-Medium, SD-3.5-Large)~\citep{esser2024scalingrectifiedflowtransformers}, PixArt models~\citep{chen2023pixartalphafasttrainingdiffusion}, FLUX-2~\citep{greenberg2025demystifyingfluxarchitecture}, HiDream-I1~\citep{cai2025hidreami1highefficientimagegenerative}, Qwen-Image~\citep{wu2025qwenimagetechnicalreport}, and Z-Image-Turbo~\citep{DBLP:journals/corr/abs-2511-22699}.
For autoregressive generation, we evaluate Infinity-8B~\citep{han2025infinityscalingbitwiseautoregressive} and GoT-R1-7B~\citep{duan2025gotr1unleashingreasoningcapability}.
We further include unified multimodal generative models including BAGEL~\citep{deng2025emergingpropertiesunifiedmultimodal}, Show-o~\citep{xie2025showosingletransformerunify}, Janus-Pro~\citep{chen2025janus}, BLIP3o~\citep{chen2025blip3ofamilyfullyopen}, and OmniGen2~\citep{wu2025omnigen2}.
For closed-source models, we evaluate FLUX-2-Pro and FLUX-Kontext-Pro~\citep{greenberg2025demystifyingfluxarchitecture}, Seedream 3.0~\citep{gao2025seedream}, Seedream 4.0~\citep{seedream2025seedream}, Ideogram v3 Turbo~\citep{ideogram2025ideogram3}, Imagen 4 and Imagen 4 Ultra~\citep{google_imagen_4}, Nano Banana and Nano Banana Pro~\citep{team2023gemini}, and GPT-Image-1 and GPT-Image-1.5~\citep{gpt4oimage,gpt_image_1_5}.

\vspace{2mm}
\noindent
\textbf{Evaluation Protocol.}
For each prompt, we generate a single image using the default inference configuration of each model.
The generated outputs are evaluated using the deterministic Script-as-a-Judge protocol introduced in Section~\ref{sec:deterministic_eval}.
A generation is considered correct only when all task-specific constraints are satisfied.
We report the accuracy for each domain as well as the overall average accuracy.
As this large-scale evaluation uses one generation per prompt, the reported results reflect single-sample performance rather than generation variance.

\vspace{2mm}
\noindent
\textbf{Evaluation Sets.}
We evaluate models on the full \dataset benchmark.
To examine the effect of scene complexity, we additionally compare matched Clean-Scene/Open-Scene pairs, where each pair preserves the same underlying mathematical requirement and differs only in scene realization and the lexical grounding needed to express it visually.

\subsection{Findings on Clean-Scene Math}

We begin by summarizing model performance in the~\emph{clean-scene} setting in \Cref{table:main_results}. We highlight the key findings below.

\begin{myfinding}
Closed-source models lead, but mathematical fidelity remains limited.
\end{myfinding}

\subsubsection{Closed-Source vs. Open-Source Models}
Closed-source models clearly lead under strict mathematical constraints.
Nano Banana Pro achieves the highest overall accuracy at $53.7\%$, followed by GPT-Image-1.5 at $39.1\%$ and GPT-Image-1 at $30.9\%$.
In contrast, most open-source models remain below $10\%$ overall accuracy; FLUX-2 is the strongest among them at $11.1\%$.
The large remaining gap, together with the best overall score only slightly above $50\%$, shows that exact mathematical constraint satisfaction remains challenging even for the strongest proprietary systems.

\begin{myfinding}
Performance gains remain highly uneven across domains.
\end{myfinding}

\subsubsection{Uneven Progress Across Mathematical Structures}
The results in Table~\ref{table:main_results} show that progress is highly domain dependent rather than uniform across mathematical skills.
GPT-Image-1.5 performs strongly on \textit{Counting} ($56.0\%$), \textit{Fractions} ($54.0\%$), and \textit{Plane Geometry} ($70.0\%$), but drops to $24.0\%$ on \textit{Angles}, $22.0\%$ on \textit{Functions}, and $20.0\%$ on \textit{Sets}.
Nano Banana Pro is substantially more balanced, reaching $48.0$--$72.0\%$ across all domains, yet its performance still varies by more than $30$ points between \textit{Plane Geometry} and \textit{Solid Geometry}.
These patterns indicate that current gains do not transfer uniformly across numerical, geometric, and symbolic forms of visual mathematics.

\begin{myfinding}
Precise mathematical rendering remains challenging.
\end{myfinding}

\subsubsection{Precise Mathematical Rendering}
Performance differences across domains reveal a clear gap between producing plausible mathematical diagrams and satisfying precise visual constraints.
\textit{Plane Geometry} is comparatively tractable: Nano Banana Pro reaches $72.0\%$, GPT-Image-1.5 $70.0\%$, GPT-Image-1 $68.0\%$, and FLUX-2-Pro $54.0\%$.
In contrast, \textit{Angles} and \textit{Functions} remain substantially more difficult for many models.
Performance remains particularly limited on Angles and Functions, where successful generation requires precise realization of angle measures, spatial configurations, and function geometry rather than merely producing semantically plausible diagrams.

\begin{myfinding}
Strong visual generation does not guarantee exact mathematical fidelity.
\end{myfinding}

\subsubsection{Exact Constraints Remain Brittle}
Even models with strong domain-specific peaks remain brittle under exact verification.
For example, GPT-Image-1 reaches $68.0\%$ on \textit{Plane Geometry} but only $16.0\%$ on \textit{Functions} and $20.0\%$ on \textit{Sets}; Nano Banana similarly reaches $64.0\%$ on \textit{Plane Geometry} but only $10.0\%$ on both \textit{Functions} and \textit{Sets}.
These large within-model gaps show that producing visually plausible mathematical content does not imply reliable satisfaction of the exact task constraints.

\begin{myfinding}
Reliable mathematical generation requires balanced capability across domains.
\end{myfinding}
\begin{figure}
    \centering
    \includegraphics[width=1\linewidth]{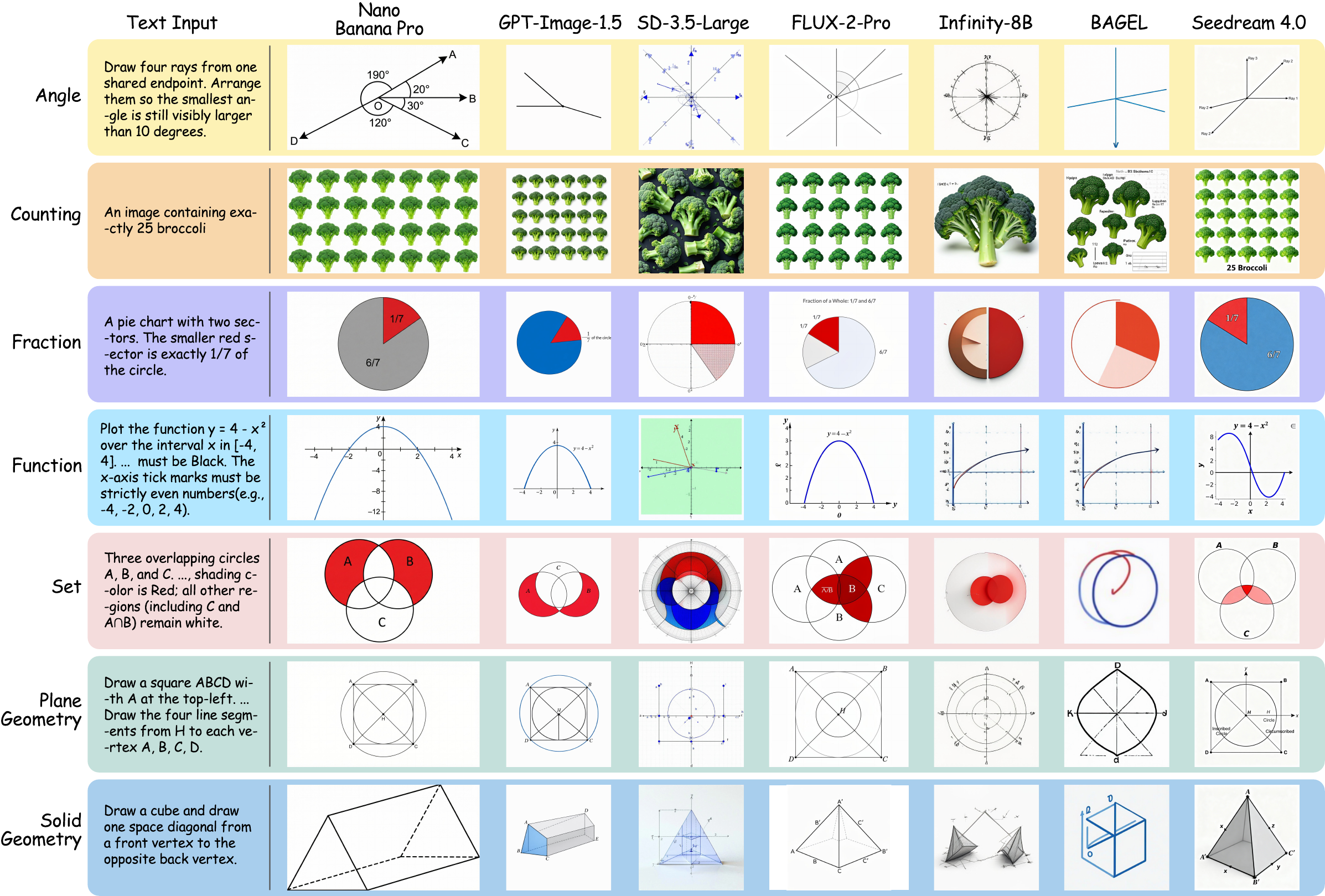}
    \caption{\small Qualitative comparison of T2I models on representative \dataset tasks.}
    \label{fig:case_study}
\end{figure}

\subsubsection{Mathematical Capability Requirements}
Across the seven domains of \dataset, the results suggest that reliable mathematical image generation requires balanced capability rather than isolated strengths.
Nano Banana Pro is the only evaluated model that reaches at least $40\%$ in every domain, yet its overall accuracy is still only $53.7\%$.
GPT-Image-1.5, despite reaching $70.0\%$ on \textit{Plane Geometry} and $56.0\%$ on \textit{Counting}, falls to $20.0$--$24.0\%$ on \textit{Sets}, \textit{Functions}, and \textit{Angles}.
These cross-domain gaps highlight the need to jointly preserve numerical consistency, geometric structure, and symbolic relations during image synthesis.

\subsection{Beyond Clean-Scene Math: Open-Scene Math}

We further study \textit{open-scene math}, where the same mathematical constraints are embedded in realistic visual environments with natural objects and background context, requiring models to preserve precise relations under more complex visual conditions.

\begin{myfinding}
Realistic scenes pose additional challenges.
\end{myfinding}
\begin{figure}[t]
    \centering
    \includegraphics[width=1\linewidth]{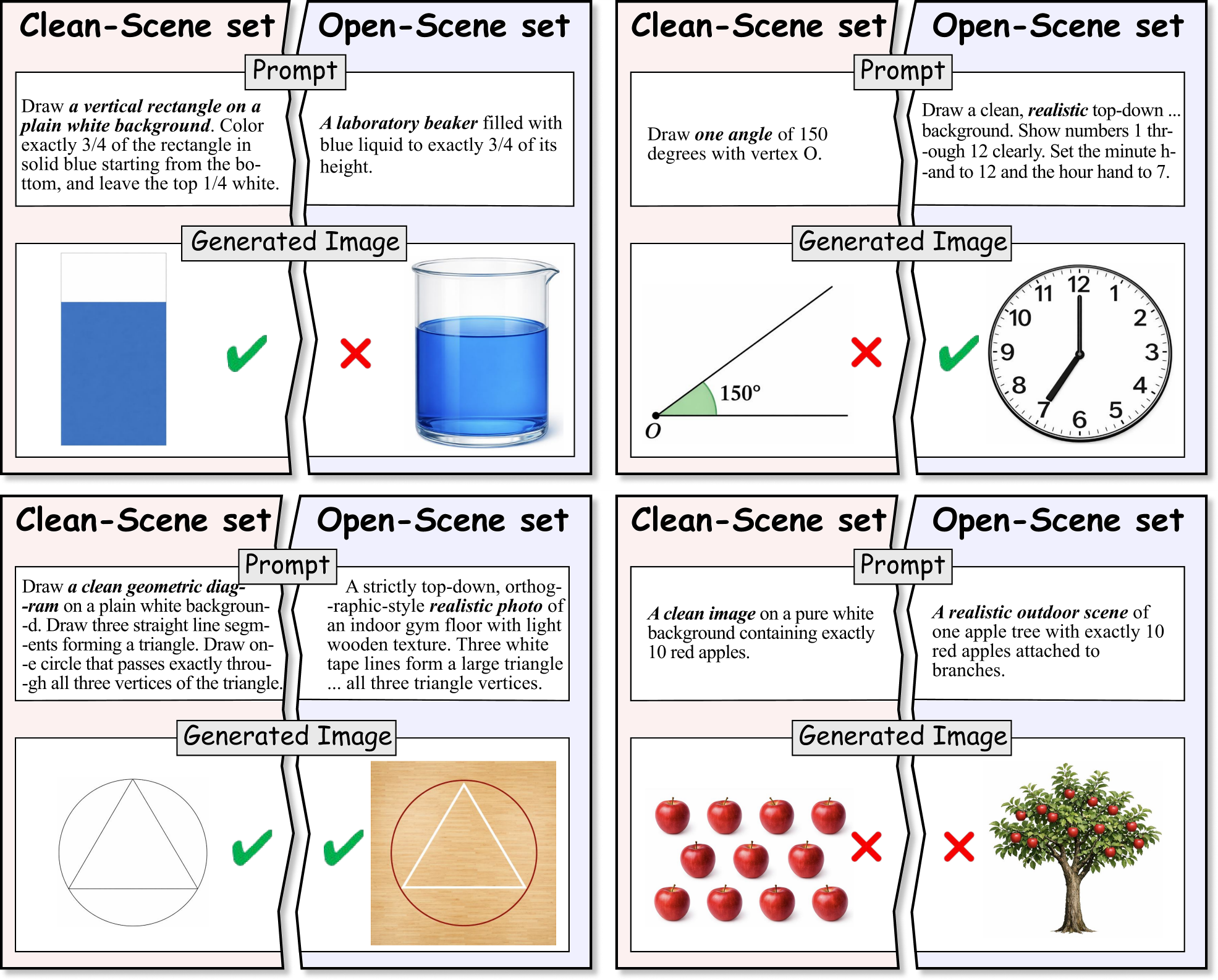}
    \caption{\small Typical success and failure examples from the~\emph{Clean-Scene} and~\emph{Open-Scene} settings.}
    \label{fig:sample_openscene}
\end{figure}

\subsubsection{Performance on Open-Scene vs. Clean-Scene}

As shown in \Cref{table:clean_open}, moving from~\emph{clean-scene} to~\emph{open-scene} generally makes exact mathematical generation harder, although the effect is not uniform across models.
Among the $30$ evaluated models, $19$ obtain lower accuracy on the Open-Scene counterparts, while $7$ improve and $4$ remain unchanged.
The degradation is particularly clear for several strong Clean-Scene models: Nano Banana Pro drops from $42.9\%$ to $32.9\%$, GPT-Image-1.5 from $38.6\%$ to $18.6\%$, FLUX-2-Pro from $30.0\%$ to $12.9\%$, and GPT-Image-1 from $32.9\%$ to $25.7\%$.
At the same time, a few weaker models improve in realistic settings, suggesting that familiar objects or scene priors can occasionally provide useful visual cues.
Overall, richer backgrounds and object interactions introduce additional variation that makes precise mathematical constraint satisfaction less reliable.
Figure~\ref{fig:sample_openscene} shows representative paired examples.

\begin{myfinding}
Clean-Scene performance remains informative but is not a substitute for Open-Scene evaluation.
\end{myfinding}

\subsubsection{Clean-Scene Performance Remains Predictive}

Despite these non-monotonic changes, Clean-Scene performance remains informative about Open-Scene capability.
Models that are strongest on the paired Clean-Scene subset generally remain among the strongest under realistic rendering: Nano Banana Pro ranks first in both settings ($42.9\%$ vs.\ $32.9\%$), while GPT-Image-1, Nano Banana, Imagen 4 Ultra, and GPT-Image-1.5 also remain near the top on Open-Scene.
However, the ranking is not preserved exactly---GPT-Image-1.5, for example, drops much more sharply than GPT-Image-1.
This indicates that Clean-Scene evaluation is a useful diagnostic of core mathematical generation ability, whereas Open-Scene evaluation additionally measures robustness to scene complexity and semantic grounding.

\begin{figure}
    \centering
    \includegraphics[width=1\linewidth]{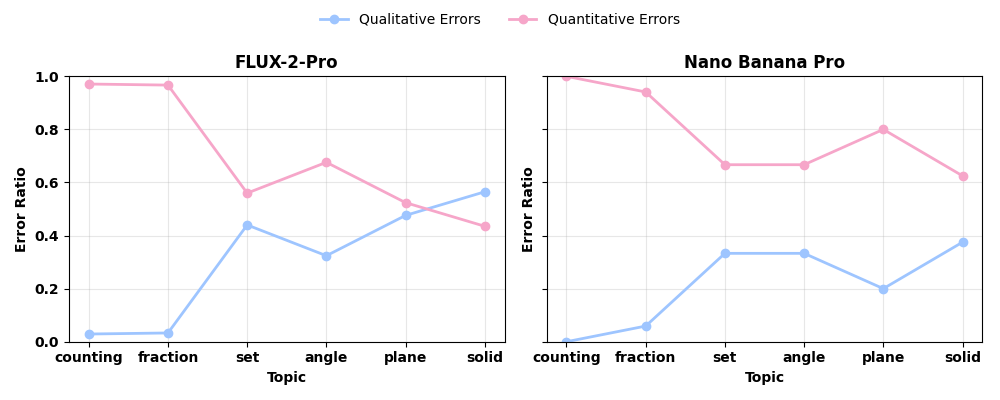}
    \caption{\small Error ratio of qualitative and quantitative failures across different mathematical topics for FLUX-2-Pro and Nano Banana Pro.}
    \label{fig:error_ratio}
\end{figure}

\subsection{Error Analysis}
\label{sec:error_analysis}

To better understand current text-to-image limitations, we manually inspected randomly sampled failures and conducted an in-depth analysis of two representative models, FLUX-2-Pro and Nano Banana Pro.
As shown in Figure~\ref{fig:error_ratio}, we group errors into qualitative failures (conceptual/structural misunderstandings) and quantitative failures (numerical/proportional deviations).
We summarize the key observations below.

\noindent
\textbf{The Evolution and Limits of Spatial Constraint Realization.}
Comparing the two models reveals a substantial reduction in qualitative errors.
Nano Banana Pro reduces qualitative errors in structurally demanding abstract domains, dropping from $10$ to $1$ in Plane Geometry and $13$ to $3$ in Solid Geometry compared to FLUX-2-Pro.
However, errors persist in logically complex tasks such as Sets, suggesting that state-of-the-art models have become more reliable at realizing basic spatial primitives but still struggle to faithfully represent complex mathematical relations.

\noindent
\textbf{The Dominance of Quantitative Bottlenecks.}
An observation across both models is the prevalence of quantitative errors, particularly in the Counting, Fractions, and Angle domains.
For instance, in the Counting task, FLUX-2-Pro and Nano Banana Pro exhibit $33$ and $17$ quantitative errors respectively, while qualitative errors remain near zero.
This contrast suggests that models can often produce the intended semantic content while remaining unreliable at precisely satisfying discrete numerical constraints.
These results motivate generation mechanisms that more explicitly represent discrete quantities, such as autoregressive layout planning or test-time verification loops, to bridge the gap between continuous image synthesis and exact mathematical constraints.
\begin{figure}
    \centering
    \includegraphics[width=1\linewidth]{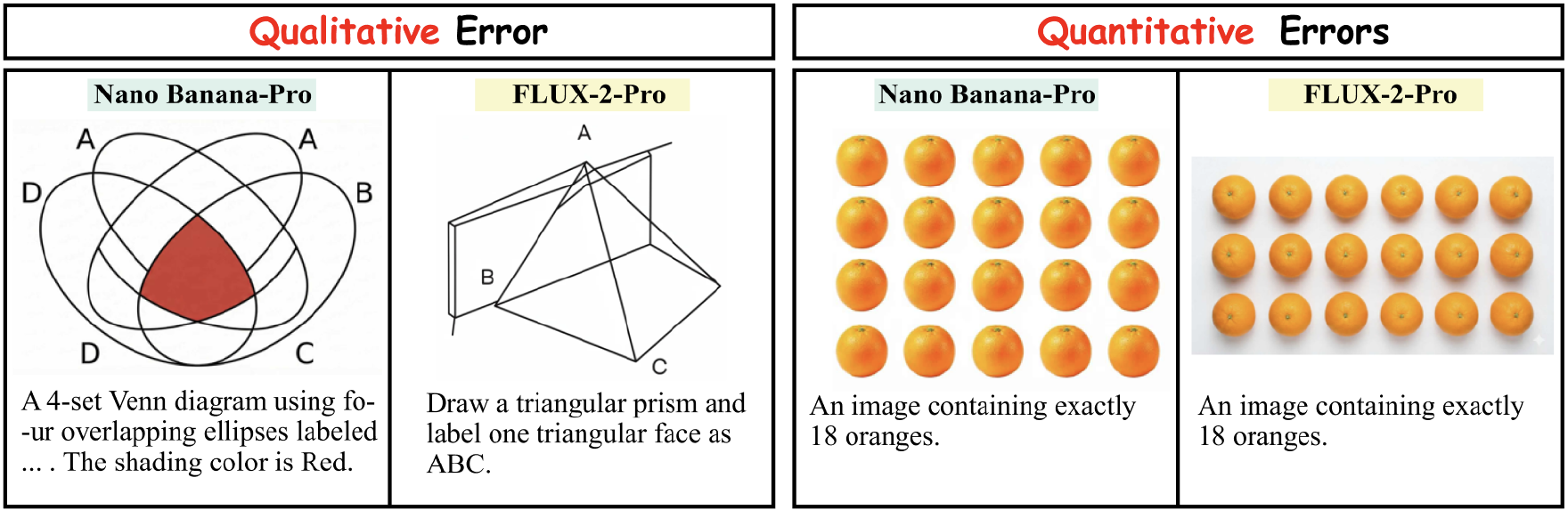}
    \caption{Representative error examples illustrating common qualitative and quantitative failures in preserving numerical and geometric constraints.}
    \label{fig:error_examples}
\end{figure}

\section{Conclusion}

In this paper, we introduced \dataset, a comprehensive benchmark designed to evaluate the mathematical generation capabilities of text-to-image models. Spanning seven core domains and $420$ problems, \dataset moves beyond subjective aesthetic evaluation toward deterministic verification of mathematical correctness via our \textit{Script-as-a-Judge} protocol.
Our extensive experiments demonstrate that while modern T2I models excel at semantic rendering, they exhibit severe deficiencies in strict mathematical understanding and reasoning, often failing to respect basic numerical counts, geometric constraints, and functional accuracy.
These findings highlight a substantial gap between visually plausible generation and constraint-faithful mathematical rendering.
We hope \dataset serves as a foundational testbed for future research, steering the community toward models that are visually creative, mathematically precise, and scientifically reliable.

\bibliographystyle{unsrtnat}
\bibliography{main}

\end{document}